\documentclass[runningheads]{llncs}
\usepackage{graphicx}
\usepackage{amsfonts}
\usepackage{amssymb}
\usepackage{amsmath}
\usepackage{color}
\usepackage{caption}
\usepackage{subcaption}
\usepackage{multirow}
\usepackage{enumitem}
\usepackage{bbding}
\begin{document}
\title{One Hyper-Initializer for All Network Architectures in Medical Image Analysis}
%
%
\author{
    Fangxin Shang\dag \and
    Yehui Yang\dag \thanks{Contact: yangyehuisw@126.com} \and
    Dalu Yang \and
    Junde Wu \and
    Xiaorong Wang \and
    Yanwu Xu
}
%
%

\institute{Intelligent Healthcare Unit, Baidu, Beijing, China \\
\dag equal contribution}

\maketitle              
%
\begin{abstract}
Pre-training is essential to deep learning model performance, especially in medical image analysis tasks where limited training data are available.
However, existing pre-training methods are inflexible as the pre-trained weights of one model cannot be reused by other network architectures. 
In this paper, we propose an architecture-irrelevant hyper-initializer, which can initialize any given network architecture well after being pre-trained for only once. 
The proposed initializer is a hypernetwork which takes a downstream architecture as input graphs and outputs the initialization parameters of the respective architecture. 
We show the effectiveness and efficiency of the hyper-initializer through extensive experimental results on multiple medical imaging modalities, especially in data-limited fields. 
Moreover, we prove that the proposed algorithm can be reused as a favorable plug-and-play initializer for any downstream architecture and task (both classification and segmentation) of the same modality. 
The code of this paper will be released later.

\end{abstract}

\section{Introduction}

Deep learning algorithms have made great progress in various medical imaging modalities, such as fundus photography \cite{googleJAMA,Yang2021RobustCL}, computed tomography(CT) \cite{DongangWang2020MaskedMN,HanxiaoZhang2020LearningWS}, X-Ray \cite{KarimHammoudi2021DeepLO}, and magnetic resonance imaging (MRI) \cite{JananiVenugopalan2021MultimodalDL}.
However, the deep models often crave a large scale of annotated data, while labeled data of medical imaging tasks are limited due to the time-consuming and expensive annotating process. 

To achieve better performance with limited data, there are mainly two kinds of weight initializing approaches for the cold-start of medical imaging tasks. One is supervised pre-training, where pre-designed deep networks are trained on a large scale labeled source data (e.g., ImageNet \cite{imagenet_cvpr09}) before fine-tuned for downstream tasks.  \cite{rajpurkar2017chexnet,LePeng2021RethinkTL,LaithAlzubaidi2021NovelTL}. The other is self-supervised pre-training, where a model is trained on a large number of unlabeled data of the same modality as the downstream task in a self-supervised manner. \cite{azizi2021big,ghesu2022selfsupervised,LiangChen2019SelfsupervisedLF}. 

Previous works on weight initialization indicate the following conclusions:
\begin{itemize}[leftmargin=*]
    \item Pre-training can enhance the performance of the downstream task in small data regime. Chen {\em et al.} \cite{LiangChen2019SelfsupervisedLF} show self-supervised pre-training can learn useful semantic features for multiple medical image modalities including CT, MR and ultrasound images.  Verified by COVID-19 classification task, Peng {\em et al.} \cite{LePeng2021RethinkTL} prove that pre-training using ImageNet benefits data-limited learning.
    
    \item Pre-training can speed up the convergence in training. Ghesu {\em et al.} \cite{ghesu2022selfsupervised} report that a well designed self-supervised initializer achieves convergence 72\% faster than no pre-training in pneumothorax classification. 
    Although pre-training with ImageNet does not necessarily for accuracy improvements due to the fundamental mismatch between the natural images and the medical images, it helps convergence compared to random initialization \cite{KaimingHe2019RethinkingIP,MaithraRaghu2019TransfusionUT}. 
    
    
\end{itemize}

Although initializing the models by self-/supervised pre-training seems to be a favorable paradigm for medical image analysis tasks, current pre-training methods are inflexible as it is time and energy consuming to repeat the pre-training procedures for every new architecture. Strubell {\em et al.}\cite{EmmaStrubell2019EnergyAP} report that training the deep learning models on abundant data consumes striking financial and environmental costs. The carbon emission to train the popular BERT model\cite{devlin2019bert} on NVIDIA V100 GPUs for 72 hours is roughly equivalent to a trans-American flight. Although take many efforts, ImageNet pre-training for ResNet-50\cite{KaimingHe2016DeepRL} take approximate 256 Tesla P100 GPUs/hour with Caffe2\cite{goyal2018accurate} and 225 Tesla P40 GPUs/hour with Tensorflow\cite{jia2018highly}. As more superior network architectures get proposed in the community, it is not a good bargain to pre-train all the new models for initialization on different downstream tasks.

To maintain the merits of pre-training while avoiding the  retraining procedures, we proposed an architecture-irrelevant initializer named \textbf{hyper-initializer}, which only needs to be pre-trained once to provide effective, efficient and scalable initial parameters for any unseen network. 
 
Inspired by \cite{ha2016hypernetworks} and \cite{knyazev2021parameter}, the proposed initializer is designed as a hypernetwork which mainly consists of three modules: 1) A {\em node embedding} module which embeds the nodes of the input networks ({\em i.e.} basic network operations such as convolution, batch-normalization, activation) into the feature space; 2) A {\em gated graph neural network (GatedGNN)} which learns the feature representation of the input architectures; 3) A {\em decoder} module with multiple linear and convolutional layers, which transforms the features from GatedGNN to the parameters for the input networks. Our hyper-initializer is trained with  unlabeled medical images in a self-supervised way, and then predict the initial parameters of any downstream model for the same image modality. 

Different from the related hypernetwork in \cite{knyazev2021parameter} which is pre-trained with a large scale of labeled dataset, we optimize the proposed hyper-initializer in a self-supervised way. Therefore, we can leverage the feature representation from the unlabled data of the similar modality with the target tasks. To the best of our knowledge, this is the first work which can initialize any unseen medical imaging models with hypernetwork. The main benefits of this paper include:

\begin{itemize}[leftmargin=*]
    \item \textbf{Effectiveness and efficiency}: By evaluating the method on multiple medical image modalities, we show that initializing the downstream models by hyper-initializer leads to performance enhancement for the target tasks in terms of both accuracy and convergence.  Additionally, by exploiting the downstream models in an interpretation way, we find that the hyper-initializer can transfer valuable feature attention from the pre-training source to the target domain. 
    
    \item \textbf{Scalability}: Instead of the conventional initialization paradigm that every new architecture need to be retrained, the proposed hyper-initializer only needs pre-trained once for one image modality then can generate effective and efficient initial parameters for any unseen architectures and tasks. Therefore, the proposed algorithm is a convenient and scalable paradigm for the development of the cold-start tasks.
\end{itemize}

\section{Method} \label{sec:method}

\noindent \textbf{Preliminary:} Given the objective function $\mathcal{L}(\cdot)$ and a N-sample training dataset  $\mathcal{D}=\{x_i,y_i\}_{i=1}^N$, the conventional optimization procedure of a given neural network $\mathcal{A}$ can be formalized as Eq.(\ref{equ:nn_optimize}):

\begin{equation}
    \operatorname*{argmin}_w \sum_{i=1}^N\mathcal{L}(\mathcal{F}(x_i;\mathcal{A},w),y_i),
\label{equ:nn_optimize}
\end{equation}

\noindent where $w$ is the weights of $\mathcal{A}$, and $\mathcal{F}(x_i;\mathcal{A},w)$ represents a forward pass of $\mathcal{A}$ with $x_i$. As the neural networks can be described by directed acyclic graph (DAG)\cite{ha2016hypernetworks}, the operations (e.g. convolutions, batch-normalizations, activations) are represented by initial node features $G^0=[g_1^0,g_2^0,...,g_{|G|}^0]$, where each $g_i^0 \in \mathbb{R}^{d^0}$ is a one-hot vector representing one of the $d^0$ operation categories. The connection of the operations(nodes) is described by a binary adjacency matrix $E\in \{0,1\}^{|G|\times|G|}$, thus $\mathcal{A}=\{G,E\}$.

\begin{figure} 
\centering
\includegraphics[width=1.0\columnwidth,trim=0 65 90 140,clip]{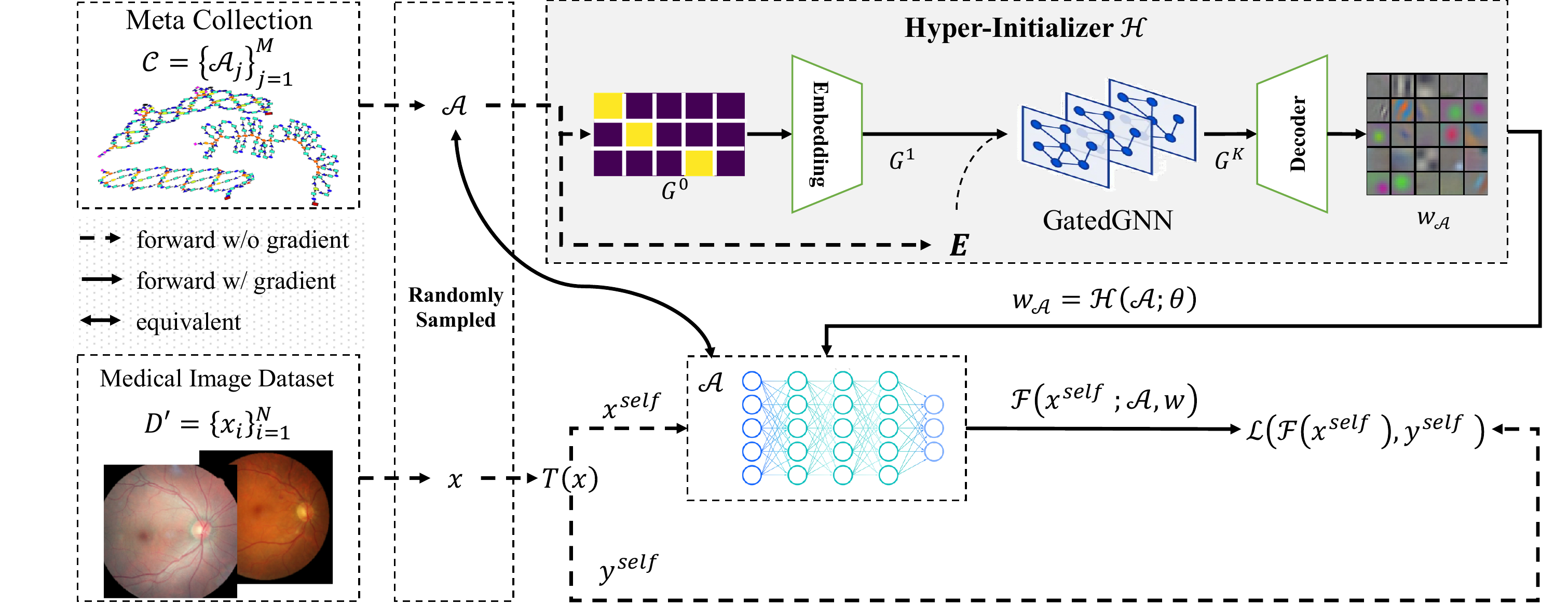}
\caption{Training procedure of the hyper-initializer. Firstly, the architecture $\mathcal{A}$ and images $x$ are randomly sampled from $\mathcal{C}$ and $\mathcal{D}'$; Secondly, the hyper-initializer $\mathcal{H}(\mathcal{A};\theta)$ generate $w_\mathcal{A}$ for the input architecture. The self-supervised samples $x^{self},y^{self}$ generated by transformation function $T(\cdot)$ are fed into $\mathcal{A}$; Finally, the self-supervised objective function is optimized to update $\theta$ by gradient descent.}
\label{fig:workflow}
\end{figure}

\noindent \textbf{Construction of the hyper-initializer:} In this paper, we propose the hyper-initializer that directly predicts modality-specific initialization parameters {\em in a single forward pass} for any neural architectures. As illustrated in the gray area of Fig.\ref{fig:workflow}, the proposed hyper-initializer $\mathcal{H}(\mathcal{A};\theta)$ mainly consists of \textbf{three} modules: 1) The {\em node embedding} module embed the initial node features $G^0$ to d-dimensional features $G^1\in \mathbb{R}^{|G|\times d}$ through an embedding layer, which is a lookup table with learnable parameters. 2) A {\em K-layer graph neural network (GatedGNN)} variant from the most recent research \cite{knyazev2021parameter} is applied to learn valuable features and associations implicit in the computational graph. The forward pass equation of GatedGNN is:

\begin{equation}
\begin{aligned}
\forall k \in [1, ..., K]: [m_v^k &= \sum_{u\in\mathcal{N}_v}{MLP(g_u^k)} + \sum_{u\in\mathcal{N}_v^{sp}} \frac{1}{s_{vu}} {MLP_{sp}(g_u^k)}], \\
g_v^{k+1} &= GRU(g_v^k,m_v^k), 
\end{aligned}
\label{equ:hypernet_fw}
\end{equation}

\noindent where $g_v^k$ indicate the features of node $v$ extracted from $k$-th graph layer; $MLP(\cdot)$ and $MLP_{sp}(\cdot)$ are two multi-layer perceptions. The $s_{vu}$ is the shortest path between node $u$ and $v$; $\mathcal{N}_v$ corresponds to the incoming neighbor of node $v$; $\mathcal{N}_v^{sp}$ are neighbors satisfying $1<s_{vu} \leq s^{max}$, where $s^{max}$ is a hyperparameter which is set to 50. $GRU(\cdot)$ is the update function of the Gated Recurrent Unit \cite{cho2014properties}. 3) A {\em decoder} module with multiple linear and convolutional layers transforms the GatedGNN output $g_v^K$ to the parameter $w_v$ for node(operation) $v$. The detailed architecture of the hyper-initializer is presented in the supplementary materials due to the limitation of paper length. 

\noindent \textbf{Hyper-initializer training:} The training procedure of the hyper-initializer is shown in Fig.\ref{fig:workflow}. Firstly, we construct the architecture collection $\mathcal{C}=\{\mathcal{A}_i\}_{i=1}^{M}$ with $M=10^6$ architectures sampled from the network design space. The space covers a high variety of stems, including VGG \cite{simonyan2014very}, ResNets \cite{KaimingHe2016DeepRL}, MobileNet \cite{howard2017mobilenets}, and 
others more recently proposed architectures. We follow the hypothesis in \cite{knyazev2021parameter} that increased training diversity can improve the hyper-initializer generalizability to unseen architectures. 
Secondly, the images $x$ and architectures $\mathcal{A}$ are randomly sampled from unlabeled medical image datasets $\mathcal{D}'=\{x_i\}$ and architecture collection $\mathcal{C}$. The transformation function $x^{self}, y^{self}=T(x)$ generates self-supervise sample pairs. In this paper, $T(x)$ is implemented by rotation angle classification proposed by \cite{gidaris2018unsupervised}. Finally, along with the conventional optimization procedure in Eq.\ref{equ:nn_optimize}, the parameters $\theta$ of the hyper-initializer $\mathcal{H}(\mathcal{A};\theta)$ can be achieved by minimizing the bi-level self-supervised optimization problem: 

\begin{equation}
    \operatorname*{argmin}_\theta \sum_{j=1}^M\sum_{i=1}^N{\mathcal{L}(\mathcal{F}(x_i^{self};\mathcal{A}_j,\mathcal{H}(\mathcal{A}_j;\theta), y_i^{self})},
\label{equ:self_meta_optimize}
\end{equation}

\noindent where the parameter set $\theta=\{\theta_{emb}, \theta_{gnn},\theta_{dec}\}$, represents the learnable parameters of embedding encoder, GatedGNN, and decoder respectively.

Once the hyper-initializer reaches convergence, the downstream tasks with any unseen architecture can fine-tune from the parameters generated by hyper-initializer in seconds, which is convenient and time-saving. 

\section{Experiments}

\textbf{Datasets:} To evaluate the proposed algorithm sufficiently, we use six publicly available medical image datasets over three modalities including fundus image, computed tomography (CT), and  X-Ray radiography. Concretely, EyePACS \cite{cuadros2009eyepacs} and APTOS2019 \cite{aptos2019}, DRIVE \cite{staal2004ridge}, and Refuge-2 \cite{li2020development} contain fundus images for classification and segmentation tasks. CT and X-Ray images are from RSNA Intracranial Hemorrhage Detection (RSNA-IHD) \cite{flanders2020construction} and RSNA Pneumonia Detection Challenge (RSNA-CXR) \cite{pan2019tackling}, respectively. The evaluation metric for diabetic retinopathy grading on EyePACS and APTOS is Kappa \cite{Yang2021RobustCL} which is a common-used metric for multi-grade classification.
For the binary classification on RSNA-IHD and RSNA-CXR datasets, we use the AUC (area under receiver operating characteristic curve) metric.
Additionally, the Dice metric is used to evaluate the performance on Refuge-2 for optic cup/disk segmentation and DRIVE for vessel segmentation. 
The details of the datasets are listed in Table \ref{tab:dataset_summary}.

\begin{table}[!]
\centering
\resizebox{0.85\textwidth}{!}{
\begin{tabular}{c|c|c|c|cc}
\hline
\multirow{2}{*}{Modality} & \multirow{2}{*}{Dataset} & \multirow{2}{*}{Task} & \multirow{2}{*}{Metric} & \multicolumn{2}{c}{Distribution} \\ \cline{5-6} 
 &  &  &  & Training Set & Evaluation Set \\ \hline
\multirow{4}{*}{Fundus} & EyePACS\cite{cuadros2009eyepacs} & {\em Cls} & Kappa & 35125 & 42670 \\
 & APTOS\cite{aptos2019} & {\em Cls} & Kappa & 2929 & 733 \\
 & Refuge-2\cite{li2020development} & {\em Seg} & Dice & 800 & 400 \\
 & DRIVE\cite{staal2004ridge} & {\em Seg} & Dice & 20 & 20 \\ \hline
CT & RSNA-IHD\cite{flanders2020construction} & {\em Cls} & AUC & 60224 & 15056 \\ \hline
X-Ray & RSNA-CXR\cite{pan2019tackling} & {\em Cls} & AUC & 24181 & 6046 \\ \hline
\end{tabular}
}
\caption{The details of the datasets in the experiments. We can see that the datasets cover multiple modalities and tasks. {\em Cls} and {\em Seg} are short for classification and segmentation tasks}
\label{tab:dataset_summary}
\end{table}

\noindent \textbf{Implementation details:} All of our experiments are implemented with PyTorch 1.10 and NVIDIA-P40. The hyper-initializer training phase last 50 epochs (iterated 50 times over $\mathcal{D}'$) with $224 \times 224$ pixels resized input images and augmented by \textit{RandomResizedCrop} with $[0.8, 1.2]$ scale range. For each modality, the unsupervised dataset $\mathcal{D}'$ consists of all images available. In the downstream fine-tune phase, all samples are resized to $512 \times 512$ pixels. Besides the data augmentation methods used for the hyper-initializer training, the \textit{RandomRotation} and \textit{RandomFlip} are also applied in the fine-tuning phase. The cosine learning rate annealing \cite{loshchilov2016sgdr} start from $0.01$ is applied in all training procedures.

\subsection{Effectiveness Evaluation}

To verify the effectiveness of the proposed algorithm, we compare the downstream models initialized by the hyper-initializer to random-initializer (train from scratch) on three modalities. To observe the sensitivity of the initializers under diffident number of training samples, we compare the two initializers under the subsets of the training data with different scales.

\begin{figure} 
    \centering
    \begin{subfigure}[b]{0.32\textwidth}
         \centering
         \includegraphics[width=\textwidth,trim=10 35 10 30,clip]{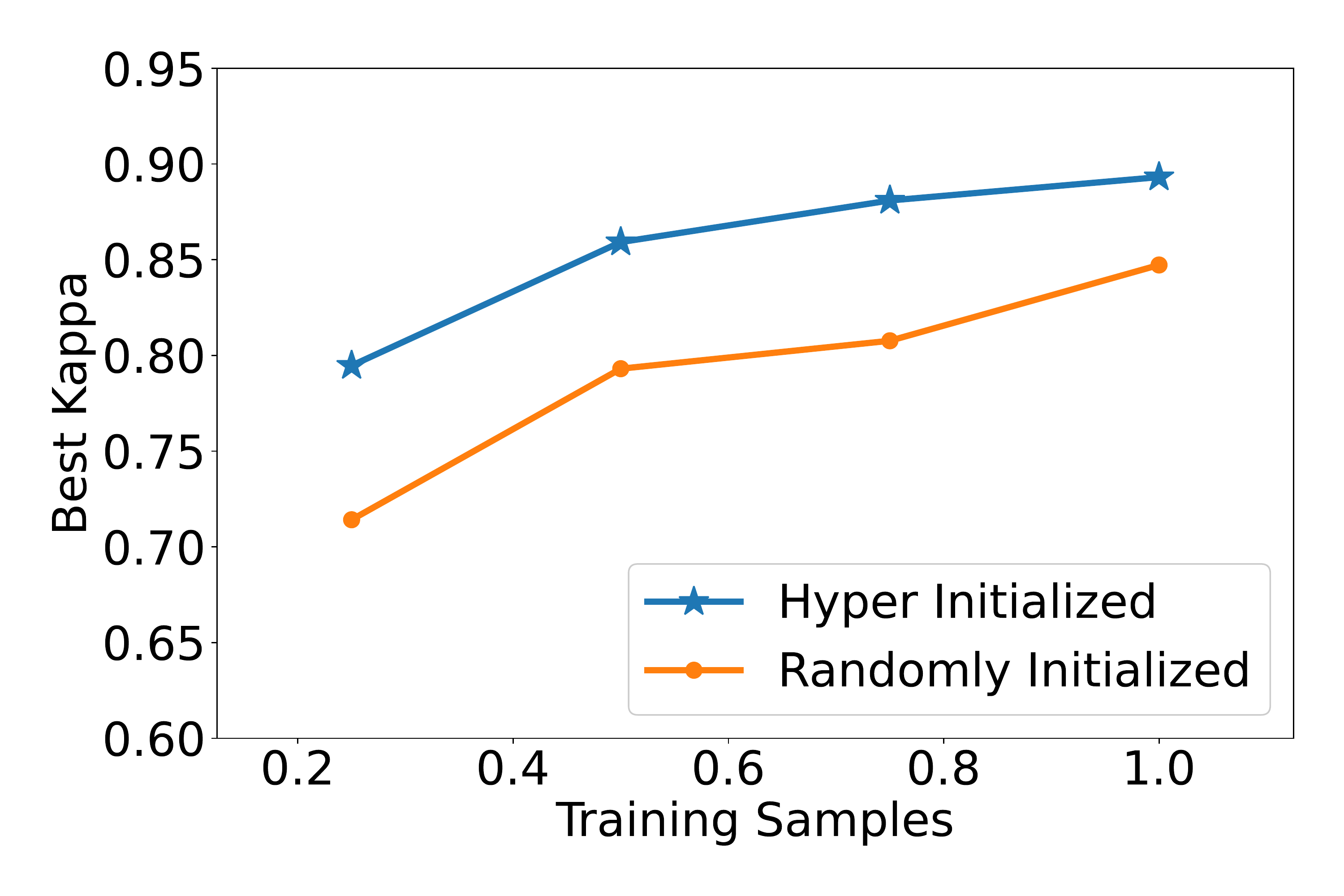}
         \caption{APTOS}
         \label{fig:effectively_by_samples_a}
    \end{subfigure}
    \begin{subfigure}[b]{0.32\textwidth}
         \centering
         \includegraphics[width=\textwidth,trim=10 35 10 30,clip]{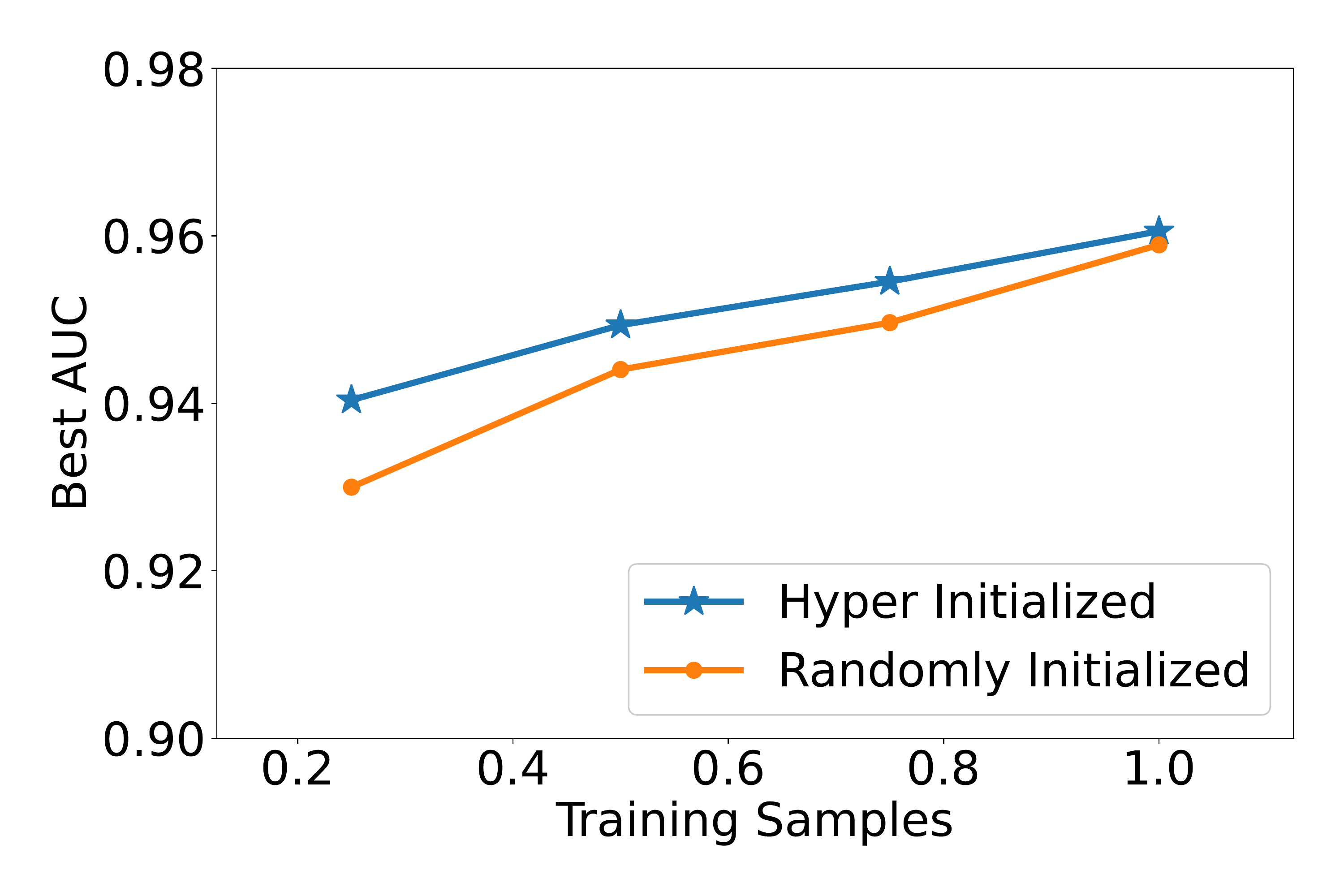}
         \caption{RSNA-IHD}
         \label{fig:effectively_by_samples_b}
    \end{subfigure}
    \begin{subfigure}[b]{0.32\textwidth}
         \centering
         \includegraphics[width=\textwidth,trim=10 35 10 30,clip]{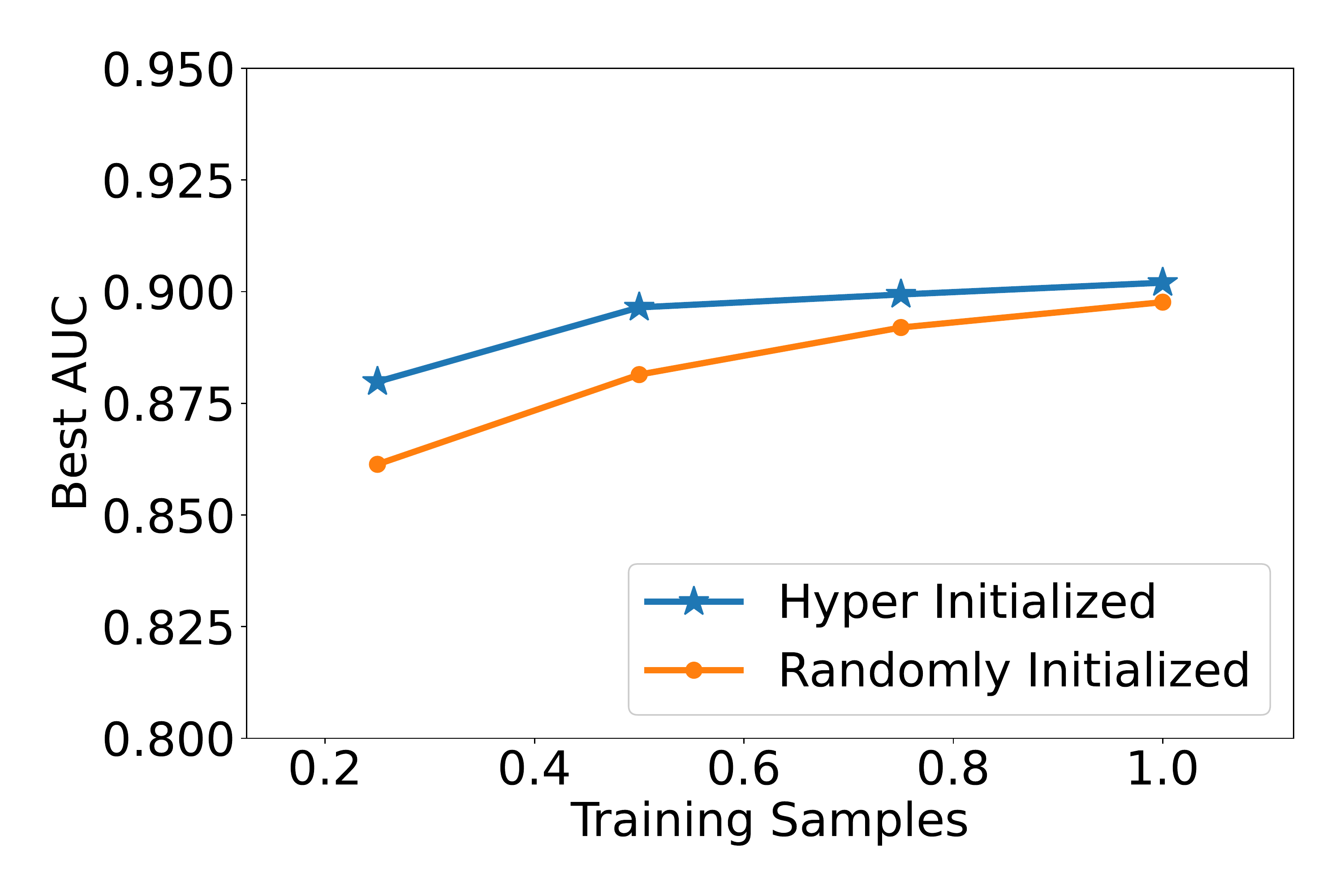}
         \caption{RSNA-CXR}
         \label{fig:effectively_by_samples_c}
    \end{subfigure}
    \caption{The performance improvement corresponding to the number of training samples. The horizontal axis represents the retained ratio of the complete training set. The vertical axis represents the best performance achieved by the model, which indicate the model benefits of hyper-initializer under different data scales.}
    \label{fig:effectively_by_samples}
\end{figure}

As shown in Fig.\ref{fig:effectively_by_samples}, the models achieves better performance on all modalities when initialized with hyper-initializer especially when the numbers of the training samples are small, which reveal that the proposed algorithm is more significant in small data regime.


\begin{figure} 
\centering
\includegraphics[width=1.0\columnwidth,trim=0 138 0 50,clip]{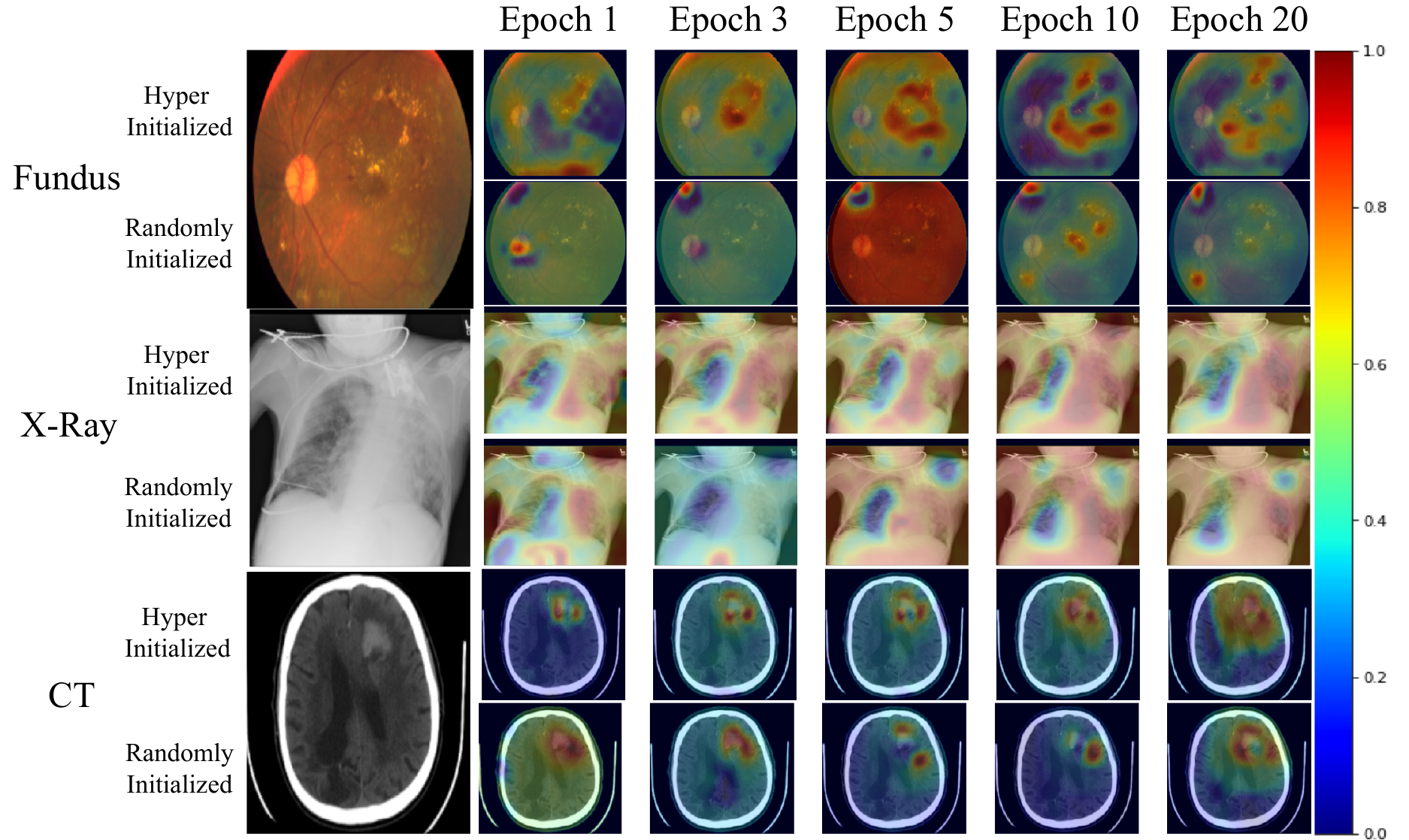}
\caption{Evolution of class activation maps (CAM) under different initialize methods. The higher values in the heatmap indicate the larger activation and attention on the corresponding regions. The hyper-initialized models quickly focus on the lesion regions of all the three modalities, while the activation regions of the random-initialized models trend to drift in the first several fine-tuning epochs.
}
\label{fig:effective_cam}
\end{figure}

Additionally, we explore the class activation maps (CAM) \cite{zhou2016learning} of the downstream model initialized by the two initialization methods, and visualize the checkpoints at epoch 1, 3, 5, 10, and 20, respectively. The CAM heatmaps in Fig.\ref{fig:effective_cam} indicate the relevant regions generated by the classification models in an interpretation way. We can see that almost all the salience regions indicated by the hyper-initialized models cover the the class-related lesions in the early fine-tuning epochs. However, the attention of the random-initialized models are various in the first several epochs. 
Therefore, the models initialized with the proposed algorithm can quickly achieve stable and accurate class-related attention, which reveal that the hyper-initializer transfer valuable modality-specific features from the pre-training source to the downstream tasks.

\subsection{Efficiency Evaluation}

\begin{figure} 
    \centering
    \includegraphics[width=1.0\columnwidth,trim=0 0 0 0,clip]{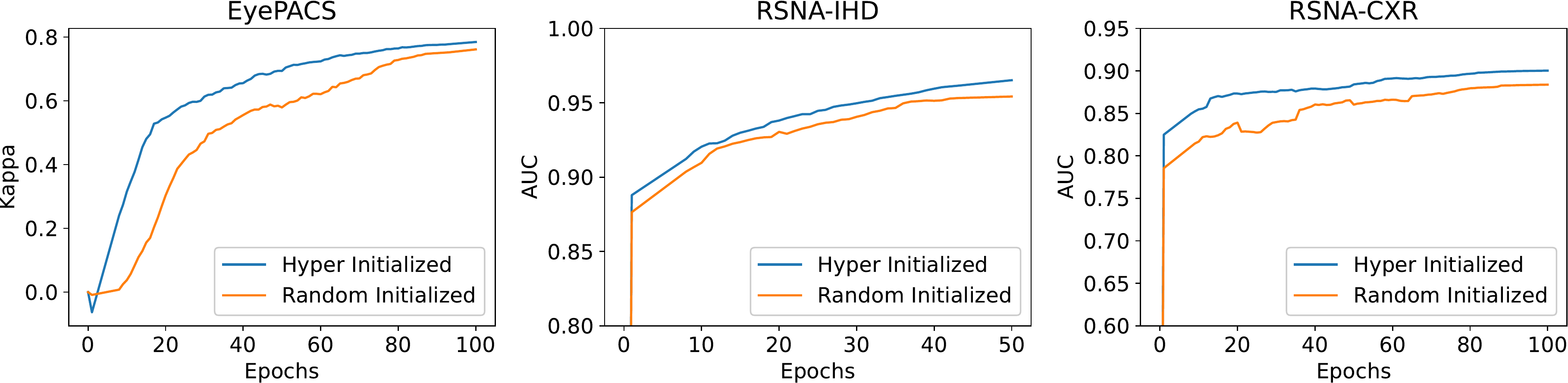}
    \caption{Convergence curve of the model on three modalities initialized by Hyper-Initializer and Random-Initializer.}
    \label{fig:efficiency_training_curve}
\end{figure}

To verify that the hyper-initializer can speed up the convergence of the downstream tasks, we record the convergence curves of the downstream tasks initialized by the hyper-initializer and random-initializer during the fine-tuning processes. As seen in Fig.\ref{fig:efficiency_training_curve}, the horizontal axis is the training epochs, and the vertical axis is the performance on the validation set corresponding to the epochs. We can see the hyper-initialized models converges faster than random-initialized ones for all three modalities. Note that the speed enhancement of hyper-initializer is inconspicuous on RSNA-IHD compared to that on other datasets. The reason is that the training samples in the RSNA-IHD (over 60 thousands images) is much lager than the other two datasets, which in line with the aforementioned conclusion that the hyper-initializer is more suitable for the limited data tasks. 

\subsection{Scalability Evaluation}

We've shown the hyper-initializer can be used for multiple image modalities in the above experiments.
In this section, we verify the scalability by applying the hyper-initializer to more neural architectures and downstream tasks.
Theoretically, since the hyper-initializer is a type of hypernetwork, it can be used for any network architectures because of the intrinsic property of the hypernetworks \cite{knyazev2021parameter}. As seen in Table \ref{tab:scale_maintable},
we show that the proposed hyper-initializer can not only generate effective initial parameters for various of ResNets \cite{KaimingHe2016DeepRL}, but also for the RegNet family \cite{radosavovic2020designing} which is the recent state-of-the-art architecture based on network architecture search. 

Besides the classification tasks, we also evaluate our method on vessel and optic cup/disk segmentation tasks which are commonly in fundus analysis. For U-Net family methods, our hyper-initializer only generates initial parameters for the encoder and leaves the decoder initialized randomly.  We can see that the proposed algorithm can also generate favorable initial parameters for the segmentation tasks. 

\begin{table}[]
\centering
\resizebox{0.9\textwidth}{!}{
\begin{tabular}{c|c|c|cc}
\hline
\multirow{2}{*}{Modality} & \multirow{2}{*}{Dataset} & \multirow{2}{*}{Architectures} & \multicolumn{2}{c}{Performance} \\ \cline{4-5} 
 &  &  & \multicolumn{1}{c|}{Random Initialize} & Hyper-Initialize \\ \hline
\multirow{7}{*}{Fundus} & \multirow{4}{*}{EyePACS} & ResNet-18 & \multicolumn{1}{c|}{0.7433} & \textbf{0.7835} \\
 &  & ResNet-50 & \multicolumn{1}{c|}{0.7870} & \textbf{0.7940} \\
 &  & ResNet-101 & \multicolumn{1}{c|}{0.7311} & \textbf{0.7734} \\
 &  & RegNetY-1.6GF & \multicolumn{1}{c|}{0.6867} & \textbf{0.7981} \\ \cline{2-5} 
 & APTOS & RegNetY-1.6GF & \multicolumn{1}{c|}{0.8472} & \textbf{0.8931} \\ \cline{2-5} 
 & Refuge-2 & UNet & \multicolumn{1}{c|}{0.7086} & \textbf{0.7816} \\ \cline{2-5} 
 & DRIVE & UNet & \multicolumn{1}{c|}{0.6853} & \textbf{0.7079} \\ \hline
\multirow{2}{*}{CT} & \multirow{2}{*}{RSNA-IHD} & ResNet-18 & \multicolumn{1}{c|}{0.9584} & \textbf{0.9639} \\
 &  & RegNetY-1.6GF & \multicolumn{1}{c|}{0.9589} & \textbf{0.9605} \\ \hline
\multirow{2}{*}{DR} & \multirow{2}{*}{RSNA-CXR} & ResNet-18 & \multicolumn{1}{c|}{0.8844} & \textbf{0.9006} \\
 &  & RegNetY-1.6GF & \multicolumn{1}{c|}{0.8976} & \textbf{0.9020} \\ \hline
\end{tabular}
}
\caption{Evaluation metrics for classification and segmentation tasks on the fundus, CT, and X-Ray. The results illustrate the proposed algorithm is scalable for various architectures and tasks.}
\label{tab:scale_maintable}
\end{table}

\section{Conclusion}

In this paper, we presented an architecture-irrelevant hyper-initializer for initializing the medical imaging models. Designed as a type of hypernetwork, the proposed initializer can generate initial parameters for any input network architectures. Optimizing the proposed algorithm with a simple self-supervised learning approach, extensive experimental evaluations reveal that the hyper-initializer can enhance the accuracy and speed up the convergence for multiple medical image modalities, especially in small data regime. To the best of our knowledge, this the the first hyper-initializer for medical imaging tasks, and we expect that the idea of hyper-initializing can be extent by the further researchers. We will try more architectures and self-supervised methods for our hyper-initializer in the future. Furthermore, we will extend the architecture collection from 2D to 3D operations, which enable the initial parameter prediction for 3D models.

%
%
%
%
\bibliographystyle{splncs04}
\bibliography{samplepaper}
\end{document}